# FUZZY SOFT ROUGH K-MEANS CLUSTERING APPROACH FOR GENE EXPRESSION DATA

Dhanalakshmi.K[1], Hannah Inbarani. H[2]

**Abstract-** Clustering is one of the widely used data mining techniques for medical diagnosis. Clustering can be considered as the most important unsupervised learning technique. Most of the clustering methods group data based on distance and few methods cluster data based on similarity. The clustering algorithms classify gene expression data into clusters and the functionally related genes are grouped together in an efficient manner. The groupings are constructed such that the degree of relationship is strong among members of the same cluster and weak among members of different clusters. In this work, we focus on a similarity relationship among genes with similar expression patterns so that a consequential and simple analytical decision can be made from the proposed Fuzzy Soft Rough K-Means algorithm. The algorithm is developed based on Fuzzy Soft sets and Rough sets. Comparative analysis of the proposed work is made with bench mark algorithms like K-Means and Rough K-Means and efficiency of the proposed algorithm is illustrated in this work by using various cluster validity measures such as DB index and Xie-Beni index.

——————————— ◆ ———————————

## 1 INTRODUCTION

### 1.1 Gene expression data

The genes are responsible for regulating the growth of cells and keeping them healthy. Normally, each cell is divided into two with control of gene. From the two cells healthy new cell take over and old ones die out. Thus the expression of a gene provides a measure of activity of a gene under certain biochemical conditions. It is known that certain diseases, such as cancer, are reflected in the change of the expression values of certain genes. In a cancer cell, mutations can "turn on" and "turn off" others in a cell.

Gene expression represents the activation level of each gene within an organism at a particular point of time. The expression value provides activity of a gene under certain biochemical conditions (also represents Samples). A gene expression data set from a microarray experiment can be represented by a real-valued expression matrix $M = \{w_{ij} \mid 1 \leq i \leq n, 1 \leq j \leq m\}$ as shown in Figure-1 where the rows represent ($G = \{\vec{g_1}, \ldots \vec{g_n}\}$) expression patterns of genes, the columns ($S = \{\vec{S_1}, \ldots \vec{S_m}\}$) represent the expression profiles of samples, and each cell $w_{ij}$ is the measured expression level of gene $i$ in sample $j$ [1]. Fig 2 represents the process flow of the gene clustering approach.

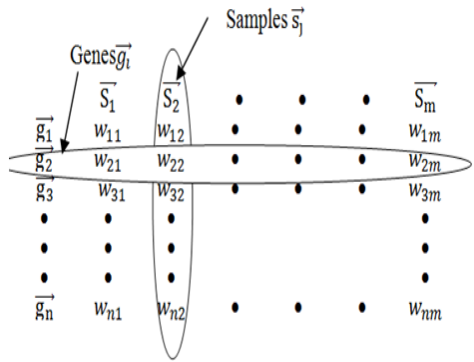

Figure 1. Gene Expression matrix

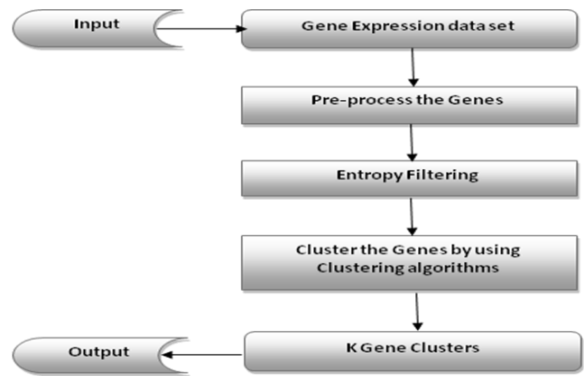

Figure 2. Methodology

• K. Dhanalakshmi M.phil Scholar, Dept. of Compuert Science, Periyar University, Tamil Nadu, India, E-mail: dhanalakshmi.src.cs@mail.com
• H. Hannah Inbarani Assistant Professor, Dept. of Computer Science, Periyar University, Tamil Nadu, India, E-mail: hhinba@gmail.com

The clustering algorithms classify gene expression data into clusters and the functionally related genes are grouped together in an efficient manner. The groupings are constructed such that the degree of relationship is strong among members of the same cluster and weak among members of different clusters. There are several clustering algorithms constructed for clustering Gene expression data such as K-means, K-medoid clustering, CLARAN'S, Fuzzy C-means, Rough K-means and so on. In this paper, Comparative analysis of the proposed work is made with two bench mark algorithms like K-Means and Rough K-Means which is explained in section

## 1.2 Fuzzy Soft Set Theory: A Review

In this part, we present the basic definitions of fuzzy soft set theory and which would be useful for subsequent discussions. Most of the definitions presented in this section may be found in [10][13][14]. The soft set theory as a new mathematical tool for dealing with uncertainties which traditional mathematical tools cannot handle [11].

**Definition 1**. Let U be an initial universe set and E be a set of parameters.
Let P(U) denote the power set of U and $A \subset E$.
   A pair (F, A) is called a soft set over U, where F is a mapping given by     F: A→ P (U). In other words, a soft set over $U$ is a parameterized family of subsets of the universe $U$. For ε Є A, F (ε) may be considered as the set of ε approximate elements of the soft set (F, A).

**Definition 2.** For two soft set (F, A) and (G, B) over a common universe U, we say that (F, A) is a soft sub set of   (G, B) if
(i). $A \subset B$, and

(ii). $\forall \varepsilon \in A, F(\varepsilon)$ and $G(\varepsilon)$ are identical approximations.
      We write $(F, A) \subset (G, B)$.
(F, A) is said to be a soft super set of (G, B),
if (G, B) is a soft subset of (F, A). We denote it by $(F, A) \supset (G, B)$.

## 1.3 Fuzzy soft set similarity

In this paper, we are often interested to know whether two patterns are identical or approximately identical or at least to what degree they are identical. Researchers have studied the problem of similarity measurement between sets; the set may be fuzzy numbers and vague sets [10][14]. P. Majumdar, S.K. Samanta has introduced the concept of similarity measure of soft set.
   Let U = {$x_1, x_2, \ldots x_n$} be the universal set of elements and E = {$e_1, e_2, \ldots, e_m$} be the universal set of parameters. Let $\hat{F} = \{F(e_i), i = 1, 2, ..., m\}$ and $\hat{G} = \{G(e_i), i = 1, 2, ..., m\}$ are two families of fuzzy soft sets. Now the similarity between $\hat{F}$ and $\hat{G}$ is found and it is denoted by S ($\hat{F}, \hat{G}$).
   Here,

$$S(\hat{F}, \hat{G}) = \max_i S_i(\hat{F}, \hat{G}),$$

$$\text{where } S_i(\hat{F}, \hat{G}) = 1 - \frac{\sum_{j=1}^{n} |\hat{F}_{ij} - \hat{G}_{ij}|}{\sum_{j=1}^{n} (\hat{F}_{ij} + \hat{G}_{ij})}$$

$$\hat{G}_{ij} = \mu_{\hat{G}(e_i)}(x_i)$$

The fuzzy soft set based similarity technique is applied to compute the similarity of genes. Only maximum similarity genes are placed in same cluster and low similarity genes are placed in different clusters.

## 2. RESEARCH MOTIVATION

Researchers are realizing that in order to achieve successful data mining, clustering algorithm is an essential component for medical diagnosis. Gene expression dataset may contain uncertain expression values. The soft set theory is a new mathematical tool for dealing with uncertainties, which traditional mathematical tools cannot handle [2]. Previous clustering algorithms so far have not used any soft set concepts and it's only applied for feature selection and classification based decision making problem. In our proposed work, we applied the Fuzzy Soft set concept for gene clustering problem. The analytical goal is to find clusters of Genes which are more similar to each other than they are to observations in different clusters.

This research area focuses on to improve the performance of gene clustering approach. To analyse which clustering algorithm gives better results among various gene expression cancer data sets, comparative analysis is made with the K-means and Rough K-means algorithms. It gives an opportunity to examine individual genes, for further medical treatments and drug discovery.

## 3. LITERATURE SURVEY

Pre-processing of gene expression data is an essential process. Various approaches for gene filtering and different types of gene clustering methods are discussed in [3]. Data clustering analysis has been extensively applied to extract information from gene expression profiles obtained with DNA microarrays [4]. Cluster analysis technique is used to identify sets of genes that are coordinately regulated. Many clustering techniques have been applied for gene expression data. For example, Limin Fu and Enzo Medico introduced a novel fuzzy clustering method for the analysis of DNA microarray data [5]. Jie Zhang, et al., have identified a set of genes that are potential biomarkers for breast cancer prognosis which can categorize the patients into two groups with distinct prognosis [6]. The authors Daxin Jiang Chun Tang Aidong Zhang, gives extensive survey for cluster analysis for gene Expression data [3].

The author S. Belciug et al. applied K-Means, SOM, and the cluster network algorithm to detect breast cancer recurrent events. The result clearly shows that the best performance was obtained by the cluster network, followed by SOM and K-Means [7]. The Author Georg Perers first introduced the Rough K-means clustering algorithm, which is applied for web mining [8].

Parvesh kumar, Siri Krishan Wasan applied both K-means and Rough K-means algorithms for Cancer data sets [9]. In this paper we have applied fuzzy soft set based similarity approach, which is taken from generalized fuzzy soft set as introduced by Pinaki Majumdar et al. [10]. Soft Set Theory was first proposed by D. Molodtsov in 1999 for dealing with uncertainties [11]. The soft sets theory can work well on the parameters that have a binary number but difficult to work with real number. To overcome this problem, Maji et al. have introduced general concepts for soft set theory, namely the fuzzy soft set, which can be used with real numbers [12], [13]. Bana Handaga and Mustafa Mat Deris used the similarity Approach on Fuzzy Soft Set based Numerical Data Classification [14]. Pinaki Majumdar et al. defined a concept generalized fuzzy soft set and several similarity measures of fuzzy soft set [15]. In This work we applied Fuzzy Soft similarity concepts in Rough K-means algorithm [16], [8] it  works well on soft set theory based classification problems, but in our proposed work we use fuzzy soft sets for clustering problem.

## 4. GENE SELECTION

One of the interesting features of microarray experiments is the fact that they group information on a large number of genes. These issues will affect biologist in many ways and we face lot of problems while go for convergence. So we must go for the dimensionality reduction of gene expression data sets. One of the characteristics of gene expression data is that it is meaningful to reduce dimension in both genes and samples, but in our thesis work we perform only the gene based clustering .So we go for pre-process the gene data's. The gene selection is also an interesting research area among gene expression data's [3]. The gene selection has three different approaches such as Filtering Approach, Wrapper Approach, and Embedded Approach.

### 4.1 Entropy filtering

In this work Entropy Filter technique is used for univariate gene selection. The effectiveness of the genes is calculated by using entropy filter method. Entropy measures the uncertainty of a random variable. For the measurement of interdependency of two random genes X and Y we used Shannon's information theory [15].

$$H(X) = -\sum_i P(X_i) \log P(X_i) \quad (1)$$

H(X) is entropy value of individual gene X. By using Entropy filter, we calculated Information Gain for the Random genes and depending upon the gain value genes may removed or selected [15].

$$IG(X,Y) = H(X) + H(Y) - H(X,Y) \quad (2)$$

### 4.2 Procedure for Entropy based Gene Selection

**Step 1**: Input gene expression matrix G = $g_1, g_2, …, g_n$ are the genes in the dataset. k = $c_1, c_2, …, c_n$ are the classes in the dataset.
**Step 2**: Calculate the Entropy for each classes using Equation (1).
**Step 3**: Calculate Information Gain for every gene by using Equation (2).
**Step 4**: Select the genes (G) with highest entropy value.

## 5. CLUSTERING ALGORITHMS

### 5.1 K-Means Clustering

The K-Means algorithm [16] is one of a group of algorithms called partitioning methods. One of the most widely used clustering techniques. It is a heuristic method where each cluster is represented by the centre of the cluster (i.e. the centroids). The K-Means algorithm is very simple and can be easily implemented in solving many problems. The K-Means algorithm is the best-known squared error-based clustering algorithm[16].The K-Means is the one of the benchmark algorithms in partitional clustering. The goal of K-Means clustering algorithm is usually to create one set of clusters that partitions the data into similar groups. Samples close to one another are assumed to be similar and the goal of the partitional clustering algorithms is to group data that are close together.

**Algorithm**

**Input:** Set of sample patterns of genes$\{X_1,X_2,…,X_m\}$, $X_i \in R^n$
**Step 1:** Choose K initial cluster centers $z_1,z_2,…,z_k$, randomly from the m patterns $\{X_1,X_2,…X_m\}$ where K<m.
**Step 2:** Assign pattern $X_i$ to cluster center $Z_j$, where I = 1, 2,…, m and j $\in$ {1,2,…,K}, If and only if $\|X_i - Z_i\| < \|X_i - Z_v\|$, p=1,2,…K and j $\neq$ p.
These are resolved arbitrarily, and compute cluster center for each point $x_i$ as follows,
$Z_i = (1/n_i)\sum X_j$, i = 1, 2, …, K.
$x_j \in Z_i$ where $n_i$ is the number of elements belonging to cluster $Z_i$.
**Step 3:** Repeat this step 2 until there are no changes in centroid values.

### 5.2 Rough K-Means Clustering

The Rough K-Means algorithm (Macqueen, 1967) is one of the group of algorithms called partitioning methods also a most widely used clustering techniques. It is a heuristic method where each cluster is represented by the centre of the cluster. Rough clustering has been useful in situations where clusters do not necessarily have crisp boundary. The Rough K- means algorithm provides a rough set theoretic flavour to the conventional rough k means algorithm to deal with uncertainty involved in cluster analysis [8].

**Algorithm**
**Input:** Data of n objects with d features, number of clusters k and values of parameters.
$W_{lower}$, $W_{upper}$ and threshold.
**Step 1:** Randomly assign each object into exactly one lower approximation $\underline{C_k}$, the objects also belongs to upper approximation $\overline{C_k}$ of the same cluster. Boundary region is $C_k^B$.
**Step 2:** Compute Cluster centroids $Z_k$.
i = 1,2…,n and j= 1,2…m
If $C_k^B = \overline{C_k} - \underline{C_k} \neq \emptyset$

$$Z_k = (W_{lower} \times \frac{\sum_{X \in \underline{C_k}} X_i}{|\underline{C_k}|}) + (W_{upper} \times \frac{\sum_{X \in \overline{C_k} - \underline{C_k}} X_i}{|\overline{C_k} - \underline{C_k}|})$$

Else
Compute new centroids,
$Z_k = \sum_{X \in C_k} X_i / |\underline{C_k}|$
**Step 3:** Find distance $d(X_i, Z_j)$
i = 1,2…,n and j= 1,2…m
**Step 4:** Find if $d(X_i, Z_j)/ d(X_i, Z_j) \leq (\varepsilon)$.
Here, $d(X_i, Z_j)$ is must a minimum one.

**Step 5:** Repeat the steps 2 – step 5 until no changes in Cluster centers.

### 5.3 Fuzzy Soft Rough K-Means Clustering

In this section, we discuss how fuzzy soft set similarity can be used to cluster cancer gene and from a set of observed samples of the gene expression values for cancer patient. The proposed Fuzzy Soft Rough K-Means algorithm is an extension of Rough K- mean but it gives effective results then Rough K-means based clustering approach [8].

**Algorithm**

1. **Pre-processing Phase**
**Step 1:** Select informative genes by using Entropy filtering approach. The Genes with low entropy value is removed.
**Step 2**: Fuzzify feature vector values using S-shaped membership function or Z-shaped membership function.

2. **Clustering Phase**
**Input:** K- No of clusters.
n- Total number of Genes.
m- Number of samples.
$W_{lower}$, $W_{upper}$ and threshold($\varepsilon$).
**Output:** K- Gene clusters.
**Step 1:** Randomly assign each object into exactly one lower Approximation $\underline{C_k}$, the objects also belongs to upper approximation $\overline{C_k}$ of the same cluster. Boundary region is $C_k^B$.
**Step 2**: Compute Cluster centroids $Z_k$.
i = 1,2…,n & j= 1,2…m, h=1, 2, …,k
If $C_k^B = \overline{C_k} - \underline{C_k} \neq \emptyset$

$$Z_k = (W_{lower} \times \frac{\sum_{X \in \underline{C_k}} X_i}{|\underline{C_k}|}) + (W_{upper} \times \frac{\sum_{X \in \overline{C_k} - \underline{C_k}} X_i}{|\overline{C_k} - \underline{C_k}|})$$

Else
Compute new centroids,
$Z_k = \sum_{X \in C_k} X_i / |\underline{C_k}|$
End
**Step 3:** Find Similarity $S_i$,
Here, $\hat{X}_i$- Represents genes, $\hat{Z}$- Represents Centriod,

$$S_i(\hat{X}, \hat{Z}) = 1 - \frac{\sum_{j=1}^{n}|\hat{x}_{ij} - \hat{z}_{hj}|}{\sum_{i=1}^{n}(\hat{x}_{ij} + \hat{z}_{hi})}$$

**Step 4:** Compute $P_i = \frac{Max(S_i)}{Min(S_i)}$ and normalize the $P_i$ values
If $P_i \geq (\varepsilon)$.
$i^{th}$ object in $K^{th}$ Cluster.
**Step 5:** Update centroids. Repeat the steps 2 – step 5, until New centroid = Old centroid.

## 6. EXPERIMENTAL RESULTS

To evaluate the performance of the various Clustering algorithms described in this paper such as Fuzzy Soft Rough K-means, Rough K-means, K-means algorithm. These algorithms are implemented for cluster genes from various gene expression data sets and the experiment is performed using MATLAB.

Cluster analysis, is an important tool in gene expression data analysis. For experimentation, we used a set of gene expression data that contains a serious of gene expression measurement of the transcript (mRNA) level of gene. In clustering gene expression data, the genes were treated as objects and the samples are treated as attributes. The cluster validity measures such as Xie-Beni index and DB-index are used as validation measures [18] [19].

### 6.1 Data set

The data set is collected from the broad institute database [17]. In the experimental analysis, Gene expression datasets of four different cancers were taken such as Breast Cancer, Leukaemia, Lung Cancer, Carcinoma Cancer. Table 1 shows the size of the original data sets and classes. After applying entropy based ranking technique for gene selection, size of the original dataset is reduced, as shown in Table 2.

TABLE 1

SUMMARY OF THE GENE EXPRESSION DATASETS

| Dataset | #Genes & Samples | #Class |
|---|---|---|
| Leukaemia Cancer | (7129 , 34) | ALL/AML |
| Lung Cancer | (7129 , 96) | Tumor/Normal |
| Carcinoma Cancer | (7457, 37) | Tumor/Normal |
| Breast Cancer | (7500, 32) | Benign/Malignant |

### 6.2 Experimental Results of Gene Filtering

Clustergram is also a one type genes expression patterns plotting method, it automatically clusters genes and samples based on gene expression patterns. So we choose Clustergram plotting for differentiate original cancer dataset and filtered data set. Figure 3 shows the Clustergram of Leukemia Cancer data set, which contains 7129 genes, 32 samples with high amount of noisy data's. After applied the Entropy based filtering techniques only informative genes are selected and noise data's also greatly removed, which is shown in fig. 4. The Filtered data set contains only 562 genes and 34 samples

TABLE 2
EXPERIMENTAL RESULTS OF ENTROPY BASED FILTERING

| Dataset | Original Datasets | Filtered Genes |
|---|---|---|
| Leukaemia Cancer | (7129, 34) | (562, 34) |
| Lung Cancer | (7129, 96) | (568, 96) |
| Carcinoma Cancer | (7457, 37) | (594, 37) |
| Breast Cancer | (7500, 32) | (570, 32) |

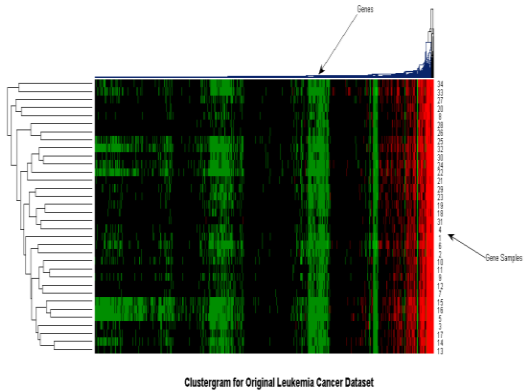

Figure 3. Cluster Dendogram Analysis of Original leukaemia Luncer Data set

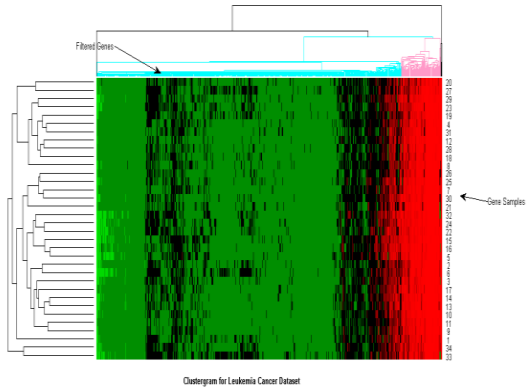

Figure 4. Cluster Dendogram Analysis of Filtered leukaemia Cancer

Fig. 5 shows the Clustergram of Lung Cancer data set, which contains 7129 genes, 96 samples with high amount of noisy data's. After applying the Entropy based filtering techniques only informative genes are selected and noise data's also greatly removed, which is shown in fig. 6. The Filtered data set contains only 568 genes and 96 samples.

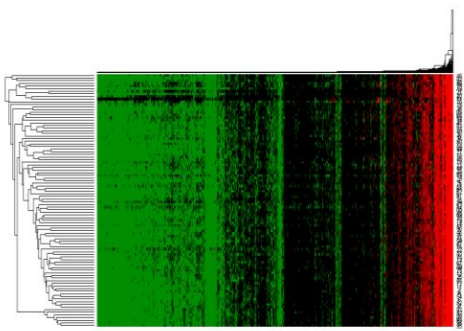

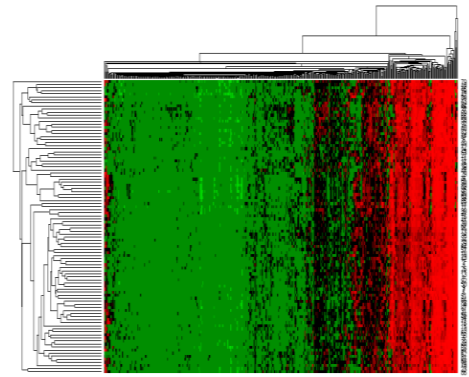

Figure 5. Cluster Dendogram Analysis of Original Lung Cancer Data set

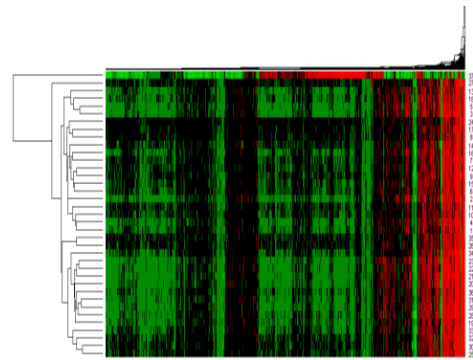

Figure 6. Cluster Dendogram Analysis of Filtered Lung Cancer Dataset

Fig. 7 shows the Clustergram of Carcinoma Cancer data set, which contains 7457 genes, 37 samples with high amount of noisy data's. After applying the Entropy based filtering techniques only informative genes are selected, which is shown in fig. 8. The Filtered data set contains only 594 genes and 37 samples with less noise.

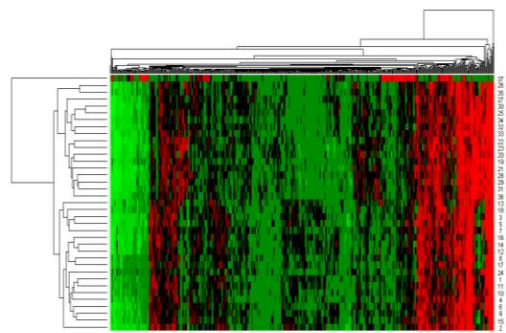

Figure 7. Cluster Dendogram Analysis of Original Gene Expression Carcinoma Cancer Data set

Figure 8. Cluster Dendogram Analysis of Filtered Gene Expression Carcinoma Cancer

Fig.9. shows the Clustergram of Breast Cancer data set, which contains 7500 genes, 32 samples with high amount of oisy data. After applying the Entropy based filtering techniques, only informative genes are selected and noise data are also greatly removed, which is shown in fig. 10. The Filtered data set contains only 570 genes and 32 samples.

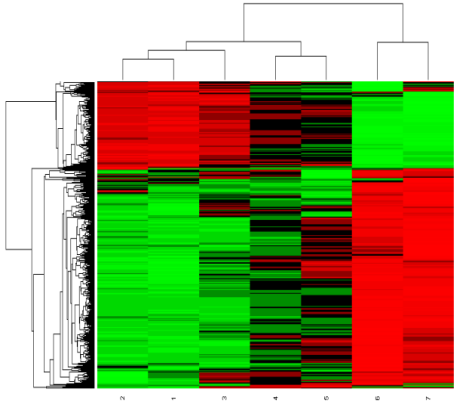

Figure 9. Cluster Dendogram Analysis of Gene Expression Original Breast Cancer Data set

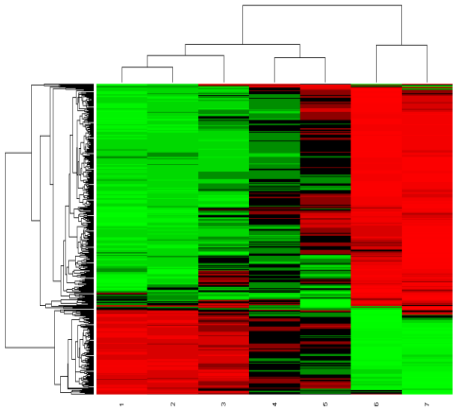

Figure 10. Cluster Dendogram Analysis of Filtered Gene Expression Breast Cancer Dataset

## 6.3 Experimental Analysis of Clustering Algorithms

In this section, the experimental results of proposed Fuzzy Soft Rough K-means algorithm are compared with rough K-means and K-means algorithms. We compare the elapsed time of three clustering algorithms, K means take less elapsed time compare to other two algorithms, Fuzzy soft Rough K-means has more internal process but presents better results in less time. Number of iterations for convergence rate is reduced with our proposed algorithm. For checking goodness of the clustering algorithms, we used DB index [17] and Xie-Beni validity Index [18]. In DB index, a good clustering algorithm has less index values; in our proposed algorithm has less DB value compare to other two clustering algorithms for all datasets (Shows in Figure 11). The Xie-Beni also less with our proposed algorithm (Figure 12). The evaluated results of cluster accuracy, Iterations for convergence are presented in Table 3, 4, 5 respectively.

TABLE 3

EXPERIMENTAL RESULTS OF K-MEANS CLUSTERING

| Data sets | DB Index | XB Index | Iterations |
|---|---|---|---|
| Leukemia | 0.1656 | 0.3572 | 6 |
| Lung | 0.1917 | 0.2056 | 5 |
| Carcinoma | 0.4672 | 0.4875 | 13 |
| Breast Cancer | 0. 4788 | 0.4902 | 7 |

TABLE 4

EXPERIMENTAL RESULTS OF ROUGH K-MEANS CLUSTERING

| Data sets | DB Index | XB Index | Iterations |
|---|---|---|---|
| Leukemia | 0.0801 | 0.3201 | 4 |
| Lung | 0.1690 | 0.2171 | 12 |
| Carcinoma | 0.4479 | 0.4776 | 15 |
| Breast | 0.4525 | 0.4952 | 12 |

TABLE 5

### EXPERIMENTAL RESULTS OF FUZZY SOFT ROUGH K-MEANS CLUSTERING

| Data sets | DB Index | XB Index | Iterations |
|---|---|---|---|
| Leukemia | 0.0673 | 0.3573 | 9 |
| Lung | 0.0154 | 0.0874 | 7 |
| Carcinoma | 0.1064 | 0.2482 | 11 |
| Breast Cancer | 0.1135 | 0.1274 | 3 |

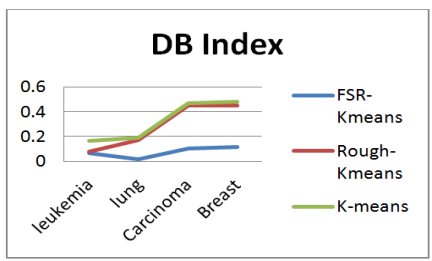

Figure 11. Performance Analysis of three Clustering algorithms measured by DB Index

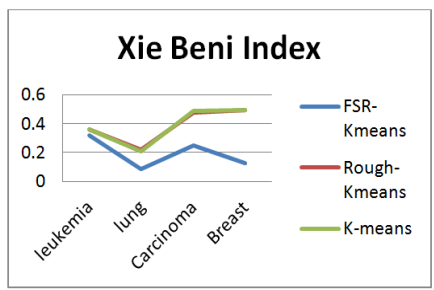

Figure 12. Performance Analysis of three clustering algorithms measured by Xie-Beni Index

## 7. CONCLUSION

Comparative analysis of the proposed work is illustrated that the proposed clustering approach provides higher accuracy than other two bench mark clustering algorithms. If the proposed clustering algorithm could integrate such partial knowledge as some clustering constraints when carrying out the clustering task, we can expect the clustering results would be more biologically meaningful. One of the characteristics of gene expression data is that it is meaningful to cluster both genes and samples, but in this thesis work we performed only the gene based clustering approach. In future, we will try to improve the performance of the clustering algorithm for both genes and samples.